\documentclass[runningheads]{llncs}
\usepackage{graphicx}
\usepackage{cite}
\usepackage{amsmath,amssymb,amsfonts}
\usepackage{algorithmic}
\usepackage{graphicx}
\usepackage{textcomp}
\usepackage{xcolor}
\usepackage{microtype}
\usepackage{times}
\usepackage{latexsym}
\usepackage{mathtools}
\usepackage{amsfonts}
\usepackage{multirow}
\usepackage{microtype}
\usepackage{pdfpages}
\usepackage{caption}
\usepackage{hyperref}
\begin{document}
\title{Document-level Relation Extraction with Cross-sentence Reasoning Graph}

%
%


\author{Hongfei Liu \inst{1} \and
Zhao Kang \inst{1} \thanks{Corresponding author.} \and
Lizong Zhang \inst{1} \and
Ling Tian \inst{1} \and
Fujun Hua \inst{2}}
\authorrunning{H. Liu et al.}
%
\institute{University of Electronic Science and Technology of China, Chengdu, China
\email{202022081525@std.uestc.edu.cn}\\
\email{\{zkang,l.zhang,lingtian\}@uestc.edu.cn} \and
Research and Development Center TROY Information Technology Co., Ltd.
Chengdu, China
\email{huafj@troy.cn}}
\maketitle              
\begin{abstract}
Relation extraction (RE) has recently moved from the sentence-level to document-level, which requires aggregating document information and using entities and mentions for reasoning. Existing works put entity nodes and mention nodes with similar representations in a document-level graph, whose complex edges may incur redundant information. Furthermore, existing studies only focus on entity-level reasoning paths without considering global interactions among entities cross-sentence. To these ends, we propose a novel document-level RE model with a \textbf{GR}aph information \textbf{A}ggregation and \textbf{C}ross-sentence \textbf{R}easoning network (GRACR). Specifically, a simplified document-level graph is constructed to model the semantic information of all mentions and sentences in a document, and an entity-level graph is designed to explore relations of long-distance cross-sentence entity pairs. Experimental results show that GRACR achieves excellent performance on two public datasets of document-level RE. It is especially effective in extracting potential relations of cross-sentence entity pairs. Our code is available at \href{https://github.com/UESTC-LHF/GRACR}{https://github.com/UESTC-LHF/GRACR}.

\end{abstract}
\section{Introduction}

\begin{figure}[t]
\centering
\includegraphics[width=0.65\textwidth,height=0.15\textwidth]{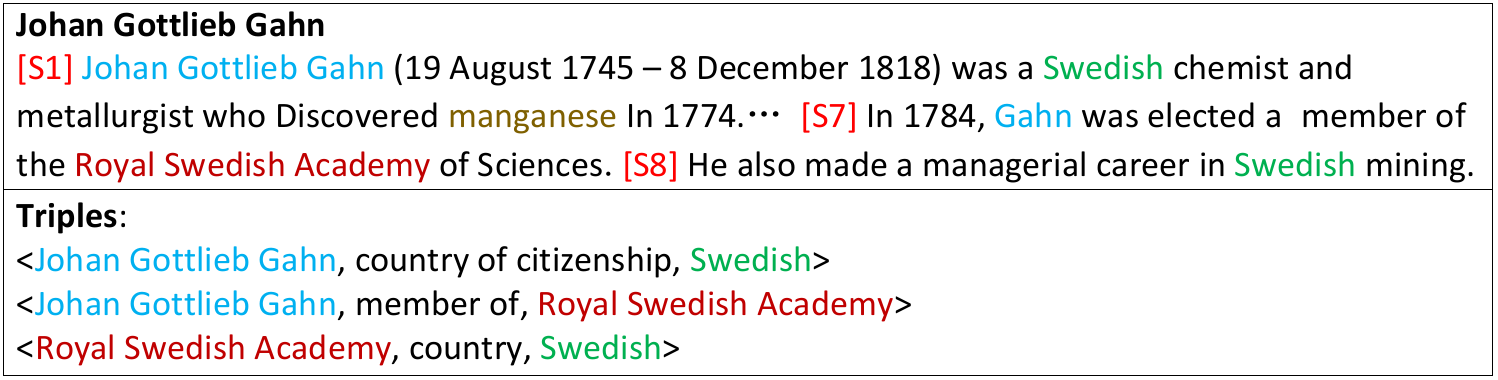}
\caption{An example of document-level RE excerpted from DocRED dataset.}
\label{example}
\end{figure}

Relation extraction (RE) is to identify the semantic relation between a pair of named entities in text.
Document-level RE requires the model to extract relations from the document and faces some intractable challenges. Firstly, a document contains multiple sentences, thus relation extraction task needs to deal with more rich and complex semantic information. Secondly, subject and object entities in the same triple may appear in different sentences, and some entities have aliase, which are often named entity mentions. Hence, the information utilized by document-level RE may not come from a single sentence. 
Thirdly, there may be interactions among different triples. Extracting the relation between two entities from different triples requires reasoning with contextual features. Figure \ref{example} shows an example from DocRED dataset \cite{yao2019docred}. It is easy to predict intra-sentence relations 
because the subject and object appear in the same sentence. However, it has a problem in identifying the inter-sentence relation between "Swedish" and "Royal Swedish Academy", whose mentions are distributed across different sentences and there exists long-distance dependencies. 

\cite{yao2019docred} proposed DocRED dataset, which contains large-scale human-annotated documents, to promote the development of sentence-level RE to document-level RE. In order to make full use of the complex semantic information of documents, recent works design document-level graph and propose models based on graph neural networks (GNN) \cite{fang2022structure}. 
\cite{christopoulou2019connecting} proposed an edge-oriented model that constructs a document-level graph with different types of nodes and edges to obtain a global representation for relation classification. \cite{nan2020reasoning} defined the document-level graph as a latent variable and induced it based on structured attention to improve the performance of document-level RE models by optimizing the structure of document-level graph. \cite{wang2020global} proposed a model that learns global representations of entities through a document-level graph, and learns local representations of entities based on their contexts. 
However, these models simply average the embeddings of mentions to obtain entity embeddings and feed them into classifiers to obtain relation labels. 
Entity and mention nodes share a similar embedding if certain entity has only one mention. Therefore, putting them in the same graph will introduce redundant information and reduce discrimination.

To address above issues, we propose a novel GNN-based document-level RE model with two graphs constructed by semantic information from the document. Our key idea is to build document-level graph and entity-level graph to fully exploit the semantic information of documents and reason about relations between entity pairs across sentences. Specifically, we solve two problems:

First, how to integrate rich semantic information of a document to obtain entity representations? We construct a document-level graph to integrate complex semantic information, which is a heterogeneous graph containing mention nodes and sentence nodes. Representations of mention nodes and sentence nodes are computed by the pre-trained language model BERT \cite{devlin2019bert}. The built document-level graph is input into the R-GCNs \cite{schlichtkrull2018modeling}, a relational graph neural network, to make nodes contain the information of their neighbor nodes. Then, representations of entities are obtained by performing logsumexp pooling operation on representations of mention nodes. In previous methods, representations of entity nodes are obtained from representations of mention nodes. 
Hence putting them in the same graph will introduce redundant information and reduce discriminability. Unlike previous document-level graph construction, our document-level graph contains only sentence nodes and mention nodes to avoid redundant information caused by repeated node representations.

Second, how to use connections between entities for reasoning? 
In this paper, we exploit connections between entities and propose an entity-level graph for reasoning. The entity-level graph is built by the positional connections between sentences and entities to make full use of cross-sentence information. It connects long-distance cross-sentence entity pairs. Through the learning of GNN, each entity node can aggregate the information of its most relevant entity nodes, which is beneficial to discover potential relations of long-distance cross-sentence entity pairs.

In summary, we propose a novel model called GRACR for document-level RE. Our main contributions are as follows:

• We propose a simplified document-level graph to integrate rich semantic information. The graph contains sentence nodes and mention nodes but not entity nodes, which avoids introducing redundant information caused by repeated node representations.

• We propose an entity-level graph for reasoning to discover potential relations of long-distance cross-sentence entity pairs. An attention mechanism is applied to fuse document embedding, aggregation, and inference information to extract relations of entity pairs.

• We conduct experiments on two public document-level relation extraction datasets. Experimental results demonstrate that our model outperforms many state-of-the-art methods.

\section{Related Work}

The research on document-level RE has a long history. 
The document-level graph provides more features for entity pairs. The relevance between entities can be captured through graph learning using GNN \cite{liu2022multilayer
}. For example, \cite{dai2022graph} utilized GNN to aggregate the neighborhood information of text graph nodes for text classification. Following this, 
\cite{christopoulou2019connecting} constructed a document-level graph with heterogeneous nodes and proposed an edge-oriented model to obtain a global representation. \cite{li2020graph} characterized the interaction between sentences and entity pairs to improve inter-sentence reasoning. \cite{zhou2020global} introduced context of entity pairs as edges between entity nodes to model semantic interactions among multiple entities. \cite{zhang2020document} constructed a dual-tier heterogeneous graph to encode the inherent structure of document and reason multi-hop relations of entities. \cite{wang2020global} learned global representations of entities through a document-level graph, and learned local representations based on their contexts. \cite{nan2020reasoning} defined the document-level graph as a latent variable to improve the performance of RE models by optimizing the structure of the document-level graph. \cite{zeng2020double} proposed a double graph-based graph aggregation and inference network (GAIN). 
Different from GAIN, our entity-level graph is a heterogeneous graph and we use R-GCNs to enable interactions between entity nodes to discover potential relations of long-distance cross-sentence entity pairs. 
\cite{wang2021document} constructed a document-level graph with rhetorical structure theory and used evidence to reasoning. \cite{shi2021document} constructed the input documents as heterogeneous graphs and utilized Graph Transformer Networks to generate semantic paths.

Unlike above document-level graph construction methods, our document-level graph contains only sentence nodes and mention nodes to avoid introducing redundant information. Moreover, previous works don't directly deal with cross-sentence entity pairs. Although entities in different sentences are indirectly connected in the graph, e.g., the minimum distance between entities across sentences is 3 and the information needs to pass through two different nodes when interacting in GLRE \cite{wang2020global}. We directly connect cross-sentence entity pairs with potential relations through bridge entities to shorten the distance of information transmission, which reduces the introduction of noise.

In addition, there are some works that try to use pre-trained models directly instead of introducing graph structures. \cite{tang2020hin} applied a hierarchical inference method to aggregate the inference information of different granularity. \cite{ye2020coreferential} captured the coreferential relations in context by a pre-training task. \cite{li2021mrn} proposed a mention-based reasoning network to capture local and global contextual information. \cite{xu2021entity} used mention dependencies to construct structured self-attention mechanism. \cite{zhou2021document} proposed adaptive thresholding and localized context pooling to solve the multi-label and multi-entity problems. These models take advantage of the multi-head attention of Transformer instead of GNN to aggregate information.

However, these studies focused on the local entity representation, which overlooks the interaction between entities distributed in different sentences \cite{luoma2020exploring}. To discover potential relations of long-distance cross-sentence entity pairs, we introduce an entity-level graph built by the positional connections between sentences and entities for reasoning.

\begin{figure*}[t]
\centering
\includegraphics[width=0.95\textwidth,height=0.44\textwidth]{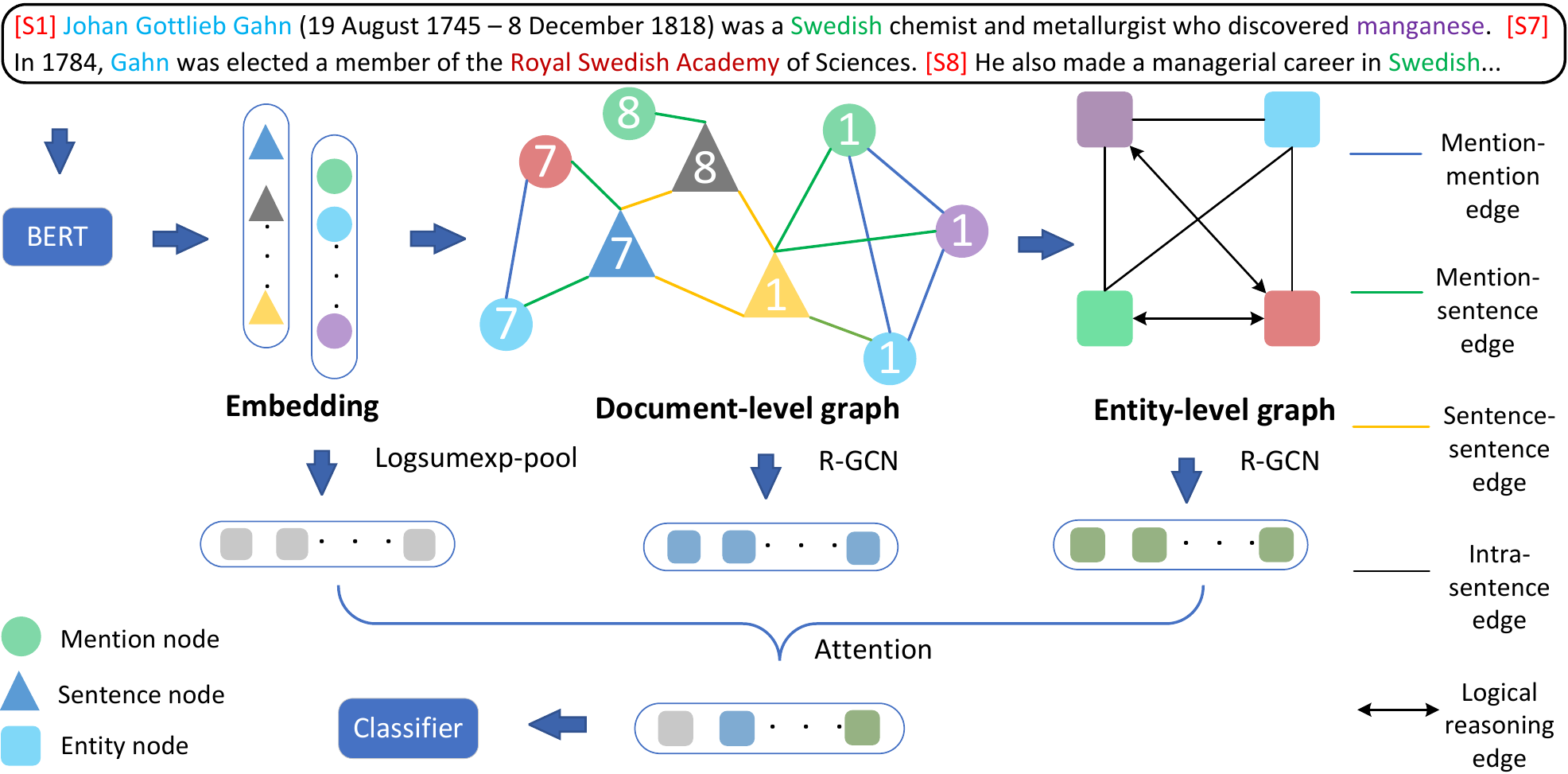}
\caption{Architecture of our proposed model. 
}
\label{model}
\end{figure*}

\section{Methodology}
In this section, we describe our proposed GRACR model that constructs a document-level graph and an entity-level graph to improve document-level RE. As shown in Figure \ref{model}, GRACR mainly consists of 4 modules: encoding module, document-level graph aggregation module, entity-level graph reasoning module, and classification module. First, in encoding module, we use a pre-trained language model such as BERT \cite{devlin2019bert} to encode the document. Next, in document-level graph aggregation module, we construct a heterogeneous graph containing mention nodes and sentence nodes to integrate rich semantic information of a document. Then, in entity-level graph reasoning module, we also propose a graph for reasoning to discover potential relations of long-distance and cross-sentence entity pairs. Finally, in classification module, we merge the context information of relation representations obtained by self-attention \cite{sorokin2017context} to make final relation prediction.

\subsection{Encoding Module}
To better capture the semantic information of document, we choose BERT as the encoder. Given an input document $\mathit{D}=\left[w_{1}, w_{2}, \ldots, w_{k}\right]$, where $w_{j} (1 \leq j \leq k)$ is the $j^{th}$ word in it. We then input the document into BERT to obtain the embeddings:
\begin{equation}\small
\mathbf{H}\!=\!\left[\mathbf{h}_{1},\mathbf{h}_{2},\ldots,\mathbf{h}_{k}\right]\!=\!\operatorname{Encoder}\!\left(\left[w_{1}, w_{2}, \ldots, w_{k}\right]\right)
\end{equation}
where $\mathbf{h}_{j} \in \mathbb{R}^{d_{w}}$ is a sequence of hidden states outputted by the last layer of BERT.

To accumulate weak signals from mention tuples, we employ logsumexp pooling \cite{jia2019document} to get the embedding $e_{i}^{h}$ of entity $\mathbf{e}_i$ as initial entity representation.
\begin{equation}\small
    \boldsymbol{e}_{i}^{h}=\log \sum_{j=1}^{N_{\mathbf{e}_i}} \exp \left(\boldsymbol{h}_{\mathbf{m}_{j}^{i}}\right)
    \label{logsumexp}
\end{equation}
where $\mathbf{m}_j^i$ is the mention $\mathbf{m}_j$ of entity $\mathbf{e}_i$, ${h}_{\mathbf{m}_j^i}$ is the embedding of $\mathbf{m}_j^i$, $N_{\mathbf{e}_{i}}$ is the number of mentions of entity $\mathbf{e}_i$ in $\mathit{D}$.

As shown in Eq.(\ref{logsumexp}), the logsumexp pooling generates an embedding for each entity by accumulating the embeddings of its all mentions across the whole document.

\subsection{Document-level Graph Aggregation Module}
To integrate rich semantic information of a document to obtain entity representations, we construct a document-level graph (\textsl{Dlg}) based on $\mathbf{H}$. 

\textsl{Dlg} has two different kinds of nodes: 

$Sentence$ $nodes$, which represent sentences in $\mathit{D}$. The representation of a sentence node $s_i$ is obtained by averaging the representations of contained words. We concatenate a node type representation $\mathbf{t}_{s} \in \mathbb{R}^{d_{t}}$ to differentiate node types. Therefore, the representations of $s_i$ is $\mathbf{h}_{s_{i}}=\left[\operatorname{avg}_{w_{j} \in s_{i}}\left(\mathbf{h}_{j}\right) ; \mathbf{t}_{s}\right]$, where $[;]$ is the concatenation operator.

$Mention$ $nodes$, which represent mentions in $\mathit{D}$. The representation of a mention node $m_i$ is achieved by averaging the representations of words that make up the mention. We concatenate a node type representation $\mathbf{t}_{m} \in \mathbb{R}^{d_{t}}$. Similar to sentence nodes, the representation of $m_i$ is $\mathbf{h}_{m_{i}}=\left[\operatorname{avg}_{w_{j} \in m_{i}}\left(\mathbf{h}_{j}\right) ; \mathbf{t}_{m}\right]$.

There are three types of edges in \textsl{Dlg}:

• Mention-mention edge. To exploit the co-occurrence dependence between mention pairs, we create a mention-mention edge. Mention nodes of two different entities are connected by mention-mention edges if their mentions co-occur in the same sentence.

• Mention-sentence edge. Mention-sentence edge is created to better capture the context information of mention. Mention node and sentence node are connected by mention-sentence edges if the mention appears in the sentence.

• Sentence-sentence edge. All sentence nodes are connected by sentence-sentence edges to eliminate the effect of sentences sequence in the document and facilitate inter-sentence interactions.

Then, we use an L-layer stacked R-GCNs \cite{schlichtkrull2018modeling} to learn the document-level graph. R-GCNs can better model heterogeneous graph that has various types of edges than GCN. Specifically, its node forward-pass update for the $(l+1)^{(th)}$ layer is defined as follows:

\begin{equation}\small
    \mathbf{n}_{i}^{l+1}\!=\!\sigma\left(\mathbf{W}_{0}^{l} \mathbf{n}_{i}^{l}\!+\!\sum_{x \in \mathit{X}} \sum_{j \in \mathit{N}_{i}^{x}} \frac{1}{\left|\mathit{N}_{i}^{x}\right|} \mathbf{W}_{x}^{l} \mathbf{n}_{j}^{l}\right)
\end{equation}
where $\sigma$(·) means the activation function, $\mathit{N}_{i}^{x}$ denotes the set of neighbors of node $i$ linked with edge $x$, and $X$ denotes the set of edge types. $\mathbf{W}_{x}^{l}, \mathbf{W}_{0}^{l} \in \mathbb{R}^{d_{n} \times d_{n}}$ are trainable parameter matrices and $d_n$ is the dimension of node representation.

We use the representations of mention nodes after graph convolution to compute the preliminary representation of entity node $e_i$ by logsumexp pooling as $e_i^{\textsl{pre}}$, which incorporates the semantic information of $e_i$ throughout the whole document. However, the information of the whole document inevitably introduce noise. We employ attention mechanism to fuse the initial embedding information and semantic information of entities to reduce noise. Specifically, we define the entity representation $e_i^{\textsl{Dlg}}$ as follows:
\begin{equation}\small
   {e}_{i}^{\textsl{Dlg}}=\operatorname{softmax}\left(\frac{\mathit{e_i^{\textsl{pre}}} \mathbf{W}_{i}^{\mathit{e_i^{\textsl{pre}}}}
   \left(e_i^{h} \mathbf{W}_{i}^{e_i^{h}}\right)^{T}}{\sqrt{d_{e_i^{h}}}}\right) e_i^{h} \mathbf{W}_{i}^{e_i^{h}}
   \label{eq4}
\end{equation}
and
\begin{equation}\small
    {e_i^{\textsl{pre}}}=\log \sum_{j=1}^{N_{\mathbf{e}_{i}}} \exp \left({n}_{m_{j}^{i}}\right)
    \label{eq5}
\end{equation}
where $\mathbf{W}_{i}^{e_i^{\textsl{pre}}}$ and $\mathbf{W}_{i}^{e_i^{h}}$ $ \in \mathbb{R}^{d_{n} \times d_{n}}$ are trainable parameter matrices. ${n}_{m_{j}^{i}}$ is mention semantic representations after graph convolution.
$d_{e_i^{h}}$ is the dimension of $e_i^{h}$.

\subsection{Entity-level Graph Reasoning Module}
To discover potential relations of long-distance cross-sentence entity pairs, we introduce an entity-level graph (\textsl{Elg}) reasoning module. \textsl{Elg} contains only one kind of node: 

$Entity$ $node$, which represents entities in $\mathit{D}$. The representation of an entity node $e_i$ is obtained from document-level graph defined by Eq. (\ref{eq5}). We concatenate a node type representation $\mathbf{t}_{e} \in \mathbb{R}^{t_{e}}$. The representations of $e_i$ is $\mathbf{h}_{e_{i}}=\left[e_i^{pre} ; \mathbf{t}_{e}\right]$.

There are two kinds of edges in \textsl{Elg}:

• Intra-sentence edge. Two different entities are connected by an intra-sentence edge if their mentions co-occur in the same sentence. For example, \textsl{Elg} uses an intra-sentence edge to connect entity nodes $e_i$ and $e_j$ if there is a path $PI_{i,j}$ denoted as $\mathbf{m}_i^{\mathbf{s}_1}$ $\rightarrow$ $\mathbf{s}_1$ $\rightarrow$ $\mathbf{m}_j^{\mathbf{s}_1}$. $\mathbf{m}_i^{\mathbf{s}_1}$ and $\mathbf{m}_j^{\mathbf{s}_1}$ are mentions of an entity pair $<$$\mathbf{e}_i$, $\mathbf{e}_j$$>$ and they appear in sentence $\mathbf{s}_1$. "$\rightarrow$" denotes one reasoning step on the reasoning path from entity node $e_i$ to $e_j$.

• Logical reasoning edge. If the mention of entity $\mathbf{e}_k$ has co-occurrence dependencies with  mentions of other two entities in different sentences, we suppose that $\mathbf{e}_k$ can be used as a bridge between entities. Two entities distributed in different sentences are connected by a logical reasoning edge if a bridge entity connects them. 
There is a logical reasoning path $PL_{i,j}$ denoted as $\mathbf{m}_i^{\mathbf{s}_1}$ $\rightarrow$ $\mathbf{s}_1$ $\rightarrow$ $\mathbf{m}_k^{\mathbf{s}_1}$ $\rightarrow$ $\mathbf{m}_k^{s_2}$ $\rightarrow$ $\mathbf{s}_2$ $\rightarrow$ $\mathbf{m}_j^{\mathbf{s}_2}$, and we apply a logical reasoning edge to connect entity nodes $e_i$ and $e_j$.

Similar to \textsl{Dlg}, we apply an L-layer stacked R-GCNs to convolute the entity-level graph to get the reasoned representation of entity $e_i^{\textsl{Elg}}$. In order to better integrate the information of entities, we employ the attention mechanism to fuse the aggregated information, the reasoned information, and the initial information of entity to form the final representation of entity. 
\begin{equation}\small
{e}_{i}^{rep}\!=\!\operatorname{softmax}\!\left(\frac{{e_i^{\textsl{Dlg}}} \mathbf{W}_{i}^{{e_i^{\textsl{Dlg}}}}\left({e_i^{\textsl{Elg}}} \mathbf{W}_{i}^{e_i^{\textsl{Elg}}}\right)^{T}}{\sqrt{d_{e_i^{Elg}}}}\right) e_i^{h} \mathbf{W}_{i}^{e_i^{h}}
\end{equation}
where $\mathbf{W}_{i}^{{e_i^{\textsl{Dlg}}}}$ and $\mathbf{W}_{i}^{{e_i^{\textsl{Elg}}}}$ $ \in \mathbb{R}^{d_{n} \times d_{n}}$ are trainable parameter matrices. $d_{e_i^{Elg}}$ is the dimension of $e_i^{Elg}$.

\subsection{Classification Module}
To classify the target relation $r$ for an entity pair $<$$e_m$, $e_n$$>$, we concatenate entity final representations and relative distance representations to represent one entity pair:
\begin{equation}\small
    \hat{{e}}_{m}=\left[{e}_{m}^{rep} ; \textsl{s}_{m n}\right],
    \hat{{e}}_{n}=\left[{e}_{n}^{rep} ; \textsl{s}_{n m}\right]
\end{equation}
where $\textsl{s}_{m n}$ denotes the embedding of relative distance from the first mention of $e_m$ to that of $e_n$ in the document. $\textsl{s}_{n m}$ is similarly defined. 

Then, we concatenate the representations of $\hat{{e}}_{m}$, $\hat{{e}}_{n}$ to form the target relation representation $\mathbf{o}_{r}=\left[\hat{{e}}_{m} ; \hat{{e}}_{n}\right]$.

Furthermore, following \cite{wang2020global}, we employ self-attention \cite{sorokin2017context} to capture context relation representations, which can help us exploit the topic information of the document:
\begin{equation}\small
\mathbf{o}_{c}=\sum_{i=1}^{p} \theta_{i} \mathbf{o}_{i}=\sum_{i=1}^{p} \frac{\exp \left(\mathbf{o}_{i} \mathbf{W} \mathbf{o}_{r}^{T}\right)}{\sum_{j=1}^{p} \exp \left(\mathbf{o}_{j} \mathbf{W} \mathbf{o}_{r}^{T}\right)} \mathbf{o}_{i}
\end{equation}
where $\mathbf{W} \in \mathbb{R}^{d_{r} \times d_{r}}$ is a trainable parameter matrix, $d_{r}$ is the dimension of target relation representations. $\mathbf{o}_{i}$ is the relation representation of the $i^{th}$ entity pair. $\theta_i$ is the attention weight for $\mathbf{o}_{i}$. $p$ is the number of entity pairs.

Finally, we use a feed-forward neural network (FFNN) on the target relation representation $o_r$ and the context relation representation $o_c$ for prediction. What's more, we transform the multi-classification problem into multiple binary classification problems, since an entity pair may have different relations. The predicted probability distribution of $r$ over the set $R$ of all relations is defined as follows:
\begin{equation}\small
    \ y_{r}=\operatorname{sigmoid}\left(\mathrm{FFNN}\left(\left[\mathbf{o}_{r}; \mathbf{o}_{c}\right]\right)\right)
\end{equation}
where $\ y_{r} \in \{0,1\}$.

We define the loss function as follows:
\begin{equation}\small
    \mathit{L}=-\sum_{r \in \mathit{R}}\left(y_{r}^{*} \log \left(y_{r}\right)+\left(1-y_{r}^{*}\right) \log \left(1-y_{r}\right)\right)
\end{equation}
where $y_{r}^{*} \in \{0,1\}$ denotes the true label of $r$. We employ Adam optimizer to optimize this loss function.

\section{Experiments and Results}

\hspace{0pt}\begin{minipage}[ht]{0.4\textwidth}
    \captionof{table}{Statistics of the datasets.}
\centering
\begin{tabular}{lcc}
\hline
\textrm{Statistics} & \textrm{DocRED} & \textrm{CDR}\\
\hline
{\# Train} & {3053} & {500}\\
{\# Dev} & {1000} & {500}\\
{\# Test} & {1000} & {500}\\ 
{\# Relations} & {97} & {2}\\ 
{Avg.\# Ents per Doc.} & {19.5} & {7.6}\\
\hline
\end{tabular}
\label{tab1}
\end{minipage}
\begin{table*}[ht]\small
\caption{Results on the development and test set of DocRED.
Some results are quoted from respective paper.}
    \centering
    \begin{tabular}{llcccc}
    \hline
\multicolumn{2}{l}{\multirow{2}*{\textbf{Model}}} & \multicolumn{2}{c}{\textbf{Dev}} & \multicolumn{2}{c}{\textbf{Test}} \\ 
\cline{3-6} & & {Ign $F_1$} & {$F_1$}  & {Ign $F_1$}  & {$F_1$} \\ \hline
\multirow{4}{*}{Sequence-based} & CNN \cite{yao2019docred}  & 41.58 & 43.45 &  40.33  &  42.26       \\
& LSTM \cite{yao2019docred}  &  48.44 &  50.68 & 47.71 & 50.07 \\
& BiLSTM \cite{yao2019docred}  & 48.87 &  50.94 & 48.78 & 51.06 \\
& Context-aware \cite{yao2019docred}  & 48.94 & 51.09 & 48.40 & 50.70\\ \hline
\multirow{4}{*}{Transformer-based} & BERT \cite{wang2019fine}  &  - &  54.16 &  -  &  53.20        \\
& HIN \cite{tang2020hin} &  54.29 &  56.31 &  53.70 & 55.60 \\
& CorefBERT \cite{ye2020coreferential} & 55.32 &  57.51 & 54.54 & 56.96 \\
& SSAN \cite{xu2021entity} & 57.03 & 59.19 & 55.84 & 58.16 \\ \hline
\multirow{6}{*}{Graph-based} & EoG \cite{christopoulou2019connecting} & 45.94 & 52.15 & 49.48 & 51.82 \\
& GEDA \cite{li2020graph}  &  54.52 & 56.16 & 53.71 & 55.74 \\
& GCGCN \cite{zhou2020global} & 55.43 & 57.35& 54.53 & 56.67 \\
& GLRE \cite{wang2020global} & - & - & 55.40& 57.40 \\ \
& DISCO \cite{wang2021document} & 55.91 & 57.78& 55.01 & 55.70 \\ \hline
Ours & GRACR & \textbf{57.85} & \textbf{59.73} & \textbf{56.47} & \textbf{58.54} \\ \hline
    \end{tabular}
    \label{tab2}
\end{table*}
\hspace{12pt}
\begin{minipage}[ht]{0.4\textwidth}
\centering
\captionof{table}{Results on CDR.}
    \begin{tabular}{p{2.5cm}ccc}
    \hline
Model & $F_1$ & intra-$F_1$ & inter-$F_1$ \\ \hline
LSR \cite{nan2020reasoning} & 64.8 & 68.9 & 53.1 \\
DHG \cite{zhang2020document} & 65.9 & 70.1 & 54.6 \\
HGNN\cite{shi2021document} & 64.4 & 69.2 & 51.2 \\
MRN \cite{li2021mrn} & 65.9 & 70.4 & 54.2 \\ \hline
GRACR & \textbf{68.8} & \textbf{73.9} & \textbf{55.8} \\ \hline
    \end{tabular}

    \label{tab3}
\end{minipage}

\hspace{0pt}\begin{minipage}[ht]{0.45\textwidth}
    \captionof{table}{Ablation study on the development set of DocRED.}
    \begin{tabular}{lcc}
    \hline
Model & Ign $F_1$ & $F_1$\\ \hline
GRACR & \textbf{57.85} & \textbf{59.73} \\
w/o both module & 57.33 & 59.16 \\
w/o reasoning module & 57.44 & 59.30 \\
w/o aggregation module & 57.61 & 59.57 \\
w/o reasoning edge & 57.52 & 59.48 \\
w/o intra-sentence edge & 57.51 & 59.46 \\
w/  previous Dlg & 57.13 & 58.97 \\ \hline
    \end{tabular}
    \label{tab4}
\end{minipage}
\hspace{5pt}
\begin{minipage}[ht]{0.45\textwidth}
\centering
\captionof{table}{Intra-$F_1$ and inter-$F_1$ results on DocRED.}
    \begin{tabular}{lcc}
\hline
Model & intra-$F_1$ & inter-$F_1$ \\ \hline
CNN \cite{yao2019docred} & 51.87 & 37.58 \\
LSTM \cite{yao2019docred}& 56.57 & 41.47 \\
BiLSTM \cite{yao2019docred}& 57.05 & 43.39 \\
Context-aware \cite{yao2019docred}& 56.74 & 42.26 \\
GEDA \cite{li2020graph} & 61.85 & 49.46 \\
LSR \cite{nan2020reasoning}& 65.26 & 52.05 \\ \hline
GRACR & \textbf{65.88} & \textbf{52.49} \\ \hline
    \end{tabular}
    \label{tab5}
\end{minipage}

\subsection{Dataset}
We evaluate our model on DocRED and CDR dataset. The dataset statistics are shown in Table \ref{tab1}. The DocRED dataset \cite{yao2019docred}, a large-scale human-annotated dataset constructed from Wikipedia, has 3,053 documents, 132,275 entities, and 56,354 relation facts in total. DocRED covers a wide variety of relations related to science, art, time, personal life, etc. The Chemical-Disease Relations (CDR) dataset \cite{li2016biocreative} is a human-annotated dataset, which is built for the BioCreative V challenge. CDR contains 1,500 PubMed abstracts about chemical and disease with 3,116 relational facts.

\subsection{Experiment Settings and Evaluation Metrics}
To implement our model, we choose uncased BERT-base \cite{devlin2019bert} as the encoder on DocRED and set the embedding dimension to 768. For CDR dataset, we pick up BioBERT-Base v1.1\cite{lee2020biobert}, which re-trained the BERT-base-cased model on biomedical corpora. 

All hyper-parameters are tuned based on the development set. Other parameters in the network are all obtained by random orthogonal initialization \cite{wang2020global} and updated during training.

For a fair comparison, we follow the same experimental settings from previous works. We apply $F_1$ and Ign $F_1$ as the evaluation metrics on DocRED. $F_1$ scores can be obtained by calculation through an online interface.
Furthermore, Ign $F_1$ means that the $F_1$ score ignores the relational facts shared by the training and development/test sets. We compare our model with three categories of models. sequence-based models use neural architectures such as CNN and bidirectional LSTM as encoder to acquire embeddings of entities. Graph-based models construct document graphs and use GNN to learn graph structures and implement inference. Instead of using document graph, transformer-based models adopt pre-trained language models to extract relation.

For CDR dataset, we use training subset to train the model. Depending on whether relation between two entities occur within one sentence or not, $F_1$ can be further split into intra-$F_1$ and inter-$F_1$ to evaluate the model’s performance on intra-sentence relations and inter-sentence relations. To make a comprehensive comparison, we also measure the corresponding $F_1$, intra-$F_1$ and inter-$F_1$ scores on development set.

\subsection{Main Results}


\textbf{Results on DocRED}. As shown in Table \ref{tab2}, our model outperforms all baseline methods on both development and test sets. Compared with graph-based models, both $F_1$ and Ign $F_1$ of our model are significantly improved. Compared to GLRE, which is the most relevant approach to our method, the performance improves 1.07\% for $F_1$ and 1.14\% for Ign $F_1$ on test set. Furthermore, compared to Transformer-based model SSAN, our method improves by 0.54\% for $F_1$ and 0.84\% for Ign $F_1$ on development set. With respect to sequence-based methods, the improvement is considerable.

\textbf{Results on CDR}. Table \ref{tab3} depicts the comparisons with state-of-the-art models on CDR. Compared to MRN \cite{li2021mrn}, the performance of our model approximately improves about 2.9\% for $F_1$, and 3.9\% for intra-$F_1$ and 1.6\% for inter-$F_1$. DHG and MRN produce similar results. In summary, these results demonstrate that our method is effective in extracting both intra-sentence relations and inter-sentence relations.

\subsection{Ablation Study}
We conduct a thorough ablation study to investigate the effectiveness of two key modules in our method: an aggregation module and an reasoning module. From Table \ref{tab4}, we can observe that all components contribute to model performance. 

(1) When the reasoning module is removed, the performance of our model on the DocRED development set for Ign $F_1$ and $F_1$ scores drops by 0.41\% and 0.43\%, respectively. Furthermore, we analyze the role of each edge in the reasoning module. $F_1$ drops by 0.23\% or 0.25\% when we remove intra-sentence edge or logical reasoning edge. Likewise, removing the aggregation module results in 0.24\% and 0.16\% drops in Ign $F_1$ and $F_1$. This phenomenon verifies the effectiveness of the aggregation module and the reasoning module.

(2) A larger drop occurs when two modules are removed. The $F_1$ score dropped from 59.73\% to 59.16\% and the Ign $F_1$ score dropped from 57.85\% to 57.33\%. This study validates that all modules work together can handle RE task more effective. 

(3) When we apply the document-level graph with entity nodes and more complex edge types like GLRE, the $F_1$ score dropped from 59.73\% to 58.97\% and the Ign $F_1$ score dropped from 57.85\% to 57.13\%. This result suggests that document-level graph containing complex and repetitive node information and edges can lead to information redundancy and degrade model performance.

\subsection{Intra- and Inter-sentence Relation Extraction}

In this subsection, we further analyze both intra- and inter-sentence RE performance on DocRED. The experimental results are listed in Table \ref{tab5}, from which we can find that GRACR outperforms the compared models in terms of intra- and inter-$F_1$. For example, our model obtains 0.62\% intra-$F_1$ and 0.44\% inter-$F_1$ gain on DocRED. The improvements suggest that 
GRACR not only considers intra-sentence relations, but also handles long-distance inter-sentence relations well.

\begin{figure}[t]
\centering
\includegraphics[width=0.6\textwidth,height=0.12\textwidth]
{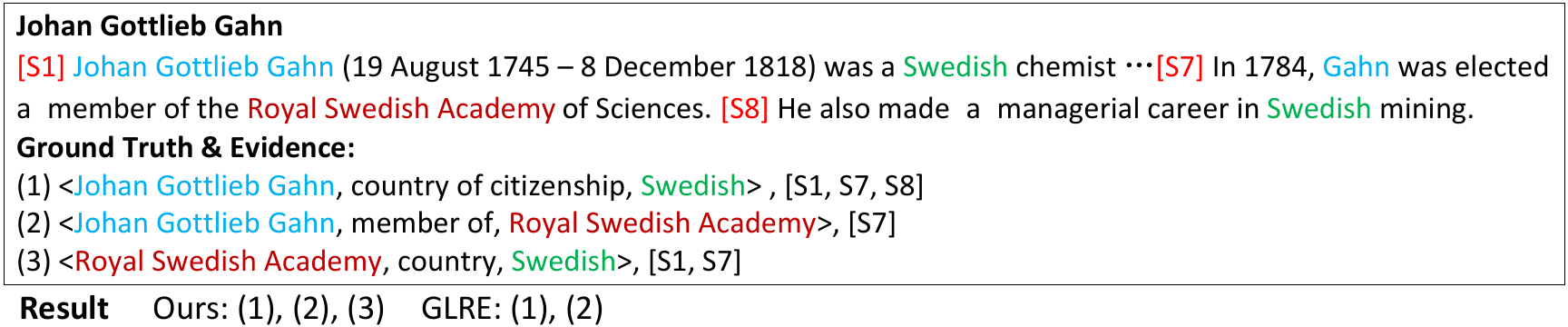}
\caption{Case study on the DocRED development set. Entities are colored accordingly.}
\label{case}
\end{figure}

\subsection{Case Study}




As shown in Figure \ref{case}, GRACR infers the relations of $<$Swedish, Royal Swedish Academy of Sciences$>$ based on the information of $S1$ and $S7$. "Swedish" and "Royal Swedish Academy of Sciences" distributed in different sentences are connected by entity-level graph because they appear in the same sentence with "Johan Gottlieb Gahn". Entity-level graph connects them together to facilitate reasoning about their relations. More importantly, our method is in line with the thinking of human logical reasoning. For example, from ground true we can know that "Gahn"'s country is "Swedish". Therefore, we can speculate that there is a high possibility that the organization he joined has a relation with "Swedish".

\section{Conclusion}

In this paper, we propose GRACR, a graph information aggregation and logical cross-sentence reasoning network, to better cope with document-level RE. GRACR applies a document-level graph and attention mechanism to model the semantic information of all mentions and sentences in a document. It also constructs an entity-level graph to utilize the interaction among different entities to reason the relations. Finally, it uses an attention mechanism to fuse document embedding, aggregation, and inference information to help identify relations. Experimental results show that our model achieves excellent performance on DocRED and CDR.

\subsubsection{Acknowledgements} This work was supported by the National Natural Science Foundation of China (Nos. 62276053, 62271125) and the Sichuan Science and
Technology Program (No. 22ZDYF3621).

%
%
%
%
\bibliographystyle{splncs04}
\bibliography{mybibliography.bib}

\begin{thebibliography}{10}
\providecommand{\url}[1]{\texttt{#1}}
\providecommand{\urlprefix}{URL }
\providecommand{\doi}[1]{https://doi.org/#1}

\bibitem{christopoulou2019connecting}
Christopoulou, F., Miwa, M., Ananiadou, S.: Connecting the dots: Document-level
  neural relation extraction with edge-oriented graphs. In: Proceedings of the
  2019 Conference on Empirical Methods in Natural Language Processing and the
  9th International Joint Conference on Natural Language Processing
  (EMNLP-IJCNLP). pp. 4925--4936 (2019)

\bibitem{dai2022graph}
Dai, Y., Shou, L., Gong, M., Xia, X., Kang, Z., Xu, Z., Jiang, D.: Graph fusion
  network for text classification. Knowledge-Based Systems  \textbf{236},
  107659 (2022)

\bibitem{devlin2019bert}
Devlin, J., Chang, M.W., Lee, K., Toutanova, K.: Bert: Pre-training of deep
  bidirectional transformers for language understanding. In: Proceedings of the
  2019 Conference of the North American Chapter of the Association for
  Computational Linguistics. pp. 4171--4186 (2019)

\bibitem{fang2022structure}
Fang, R., Wen, L., Kang, Z., Liu, J.: Structure-preserving graph representation
  learning. In: ICDM (2022)

\bibitem{jia2019document}
Jia, R., Wong, C., Poon, H.: Document-level n-ary relation extraction with
  multiscale representation learning. In: Proceedings of the 2019 Conference of
  the North American Chapter of the Association for Computational Linguistics:
  Human Language Technologies, Volume 1 (Long and Short Papers). pp. 3693--3704
  (2019)

\bibitem{lee2020biobert}
Lee, J., Yoon, W., Kim, S., Kim, D., Kim, S., So, C.H., Kang, J.: Biobert: a
  pre-trained biomedical language representation model for biomedical text
  mining. Bioinformatics  \textbf{36}(4),  1234--1240 (2020)

\bibitem{li2020graph}
Li, B., Ye, W., Sheng, Z., Xie, R., Xi, X., Zhang, S.: Graph enhanced dual
  attention network for document-level relation extraction. In: Proceedings of
  the 28th International Conference on Computational Linguistics. pp.
  1551--1560 (2020)

\bibitem{li2016biocreative}
Li, J., Sun, Y., Johnson, R.J., Sciaky, D., Wei, C.H., Leaman, R., Davis, A.P.,
  Mattingly, C.J., Wiegers, T.C., Lu, Z.: Biocreative v cdr task corpus: a
  resource for chemical disease relation extraction. Database  \textbf{1}, ~10
  (2016)

\bibitem{li2021mrn}
Li, J., Xu, K., Li, F., Fei, H., Ren, Y., Ji, D.: Mrn: A locally and globally
  mention-based reasoning network for document-level relation extraction. In:
  Findings of the Association for Computational Linguistics: ACL-IJCNLP 2021.
  pp. 1359--1370 (2021)

\bibitem{liu2022multilayer}
Liu, L., Kang, Z., Ruan, J., He, X.: Multilayer graph contrastive clustering
  network. Information Sciences  \textbf{613},  256--267 (2022)

\bibitem{luoma2020exploring}
Luoma, J., Pyysalo, S.: Exploring cross-sentence contexts for named entity
  recognition with bert. In: Proceedings of the 28th International Conference
  on Computational Linguistics. pp. 904--914 (2020)

\bibitem{nan2020reasoning}
Nan, G., Guo, Z., Sekuli{\'c}, I., Lu, W.: Reasoning with latent structure
  refinement for document-level relation extraction. In: Proceedings of the
  58th Annual Meeting of the Association for Computational Linguistics. pp.
  1546--1557 (2020)

\bibitem{schlichtkrull2018modeling}
Schlichtkrull, M., Kipf, T.N., Bloem, P., Berg, R.v.d., Titov, I., Welling, M.:
  Modeling relational data with graph convolutional networks. In: European
  semantic web conference. pp. 593--607. Springer (2018)

\bibitem{shi2021document}
Shi, Y., Xiao, Y., Quan, P., Lei, M., Niu, L.: Document-level relation
  extraction via graph transformer networks and temporal convolutional
  networks. Pattern Recognition Letters  \textbf{149},  150--156 (2021)

\bibitem{sorokin2017context}
Sorokin, D., Gurevych, I.: Context-aware representations for knowledge base
  relation extraction. In: Proceedings of the 2017 Conference on Empirical
  Methods in Natural Language Processing. pp. 1784--1789 (2017)

\bibitem{tang2020hin}
Tang, H., Cao, Y., Zhang, Z., Cao, J., Fang, F., Wang, S., Yin, P.: Hin:
  Hierarchical inference network for document-level relation extraction. In:
  Pacific-Asia Conference on Knowledge Discovery and Data Mining. pp. 197--209.
  Springer (2020)

\bibitem{wang2020global}
Wang, D., Hu, W., Cao, E., Sun, W.: Global-to-local neural networks for
  document-level relation extraction. In: Proceedings of the 2020 Conference on
  Empirical Methods in Natural Language Processing (EMNLP). pp. 3711--3721
  (2020)

\bibitem{wang2021document}
Wang, H., Qin, K., Lu, G., Yin, J., Zakari, R.Y., Owusu, J.W.: Document-level
  relation extraction using evidence reasoning on rst-graph. Knowledge-Based
  Systems  \textbf{228},  107274 (2021)

\bibitem{wang2019fine}
Wang, H., Focke, C., Sylvester, R., Mishra, N., Wang, W.: Fine-tune bert for
  docred with two-step process. arXiv preprint arXiv:1909.11898  (2019)

\bibitem{xu2021entity}
Xu, B., Wang, Q., Lyu, Y., Zhu, Y., Mao, Z.: Entity structure within and
  throughout: Modeling mention dependencies for document-level relation
  extraction. In: Proceedings of the AAAI conference on artificial
  intelligence. vol.~35, pp. 14149--14157 (2021)

\bibitem{yao2019docred}
Yao, Y., Ye, D., Li, P., Han, X., Lin, Y., Liu, Z., Liu, Z., Huang, L., Zhou,
  J., Sun, M.: Docred: A large-scale document-level relation extraction
  dataset. In: Proceedings of the 57th Annual Meeting of the Association for
  Computational Linguistics. pp. 764--777 (2019)

\bibitem{ye2020coreferential}
Ye, D., Lin, Y., Du, J., Liu, Z., Li, P., Sun, M., Liu, Z.: Coreferential
  reasoning learning for language representation. In: Proceedings of the 2020
  Conference on Empirical Methods in Natural Language Processing (EMNLP). pp.
  7170--7186 (2020)

\bibitem{zeng2020double}
Zeng, S., Xu, R., Chang, B., Li, L.: Double graph based reasoning for
  document-level relation extraction. In: Proceedings of the 2020 Conference on
  Empirical Methods in Natural Language Processing (EMNLP). pp. 1630--1640
  (2020)

\bibitem{zhang2020document}
Zhang, Z., Yu, B., Shu, X., Liu, T., Tang, H., Yubin, W., Guo, L.:
  Document-level relation extraction with dual-tier heterogeneous graph. In:
  Proceedings of the 28th International Conference on Computational
  Linguistics. pp. 1630--1641 (2020)

\bibitem{zhou2020global}
Zhou, H., Xu, Y., Yao, W., Liu, Z., Lang, C., Jiang, H.: Global
  context-enhanced graph convolutional networks for document-level relation
  extraction. In: Proceedings of the 28th International Conference on
  Computational Linguistics. pp. 5259--5270 (2020)

\bibitem{zhou2021document}
Zhou, W., Huang, K., Ma, T., Huang, J.: Document-level relation extraction with
  adaptive thresholding and localized context pooling. In: Proceedings of the
  AAAI conference on artificial intelligence. vol.~35, pp. 14612--14620 (2021)

\end{thebibliography}
\end{document}